# Generative Adversarial Synthesis of Radar Point Cloud Scenes

Muhammad Saad Nawaz[a,c], Thomas Dallmann[b], Torsten Schön[c], Dirk Heberling[a]
[a] Institute of High Frequency Technology, RWTH Aachen University, 52056 Aachen, Germany
[b] Radio Technologies for Automated and Connected Vehicles Research Group,
Technische Universität Ilmenau, Helmholtzplatz 2, 98693 Ilmenau, Germany
[c] AImotion Bavaria, Technische Hochschule Ingolstadt, Esplanade 10, 85049 Ingolstadt, Germany

## Abstract

For the validation and verification of automotive radars, datasets of realistic traffic scenarios are required, which, however, are laborious to acquire. In this paper, we introduce radar scene synthesis using GANs as an alternative to the real dataset acquisition and simulation-based approaches. We train a PointNet++ based GAN model to generate realistic radar point cloud scenes and use a binary classifier to evaluate the performance of scenes generated using this model against a test set of real scenes. We demonstrate that our GAN model achieves similar performance (~87%) to the real scenes test set.

## 1 Introduction

Radar sensors play an important role in driver assistance systems. However, their role in SAE L3 and beyond is limited to providing backup to the physical limitations of camera sensors up to now, for example, in adverse weather conditions or night situations. During the last few years, the challenges of highly automated and fully autonomous driving have demanded the availability of redundant environment perception paths parallel to the camera sensors. This has highlighted the limitations of radar-based perception algorithms, like slow processing and limited object detection and classification capabilities. Researchers have recently addressed these limitations by extending the applications of neural network-based detection and classification techniques from the camera domain into the radar domain [1] or even on fused radar-camera input [2].

However, since radar is an active sensor, the difference in inherent hardware characteristics between different radar products causes a domain gap between the acquired data of the two products. The differences include characteristics like the transmit chirp waveform as well as the design of the antenna array resulting in different transmit and receive antenna patterns. This domain gap makes the re-usability of radar data for learning across different radar products an extremely difficult and computationally expensive task. Additionally, since radar data is quite non-intuitive for visual observation compared to camera data, it is hard to label the acquired data for ground-truth generation without any synchronized reference camera. Due to these challenges, training neural networks for radar perception tasks becomes a costly commodity.

One possible solution to avoid expensive radar data collection and manual labeling is to use simulation methods like ray tracing [3]. These simulation methods provide an inexpensive way of getting synthetic training data; however, since such methods can only approximate reality under the constraint of available simulation resources, they generally fail to simulate the clutter and other target fluctuations affecting the data in real environments. Hence neural networks trained on data generated using such methods do not capture the variations of the real environment and perform poorly during inference on real data.

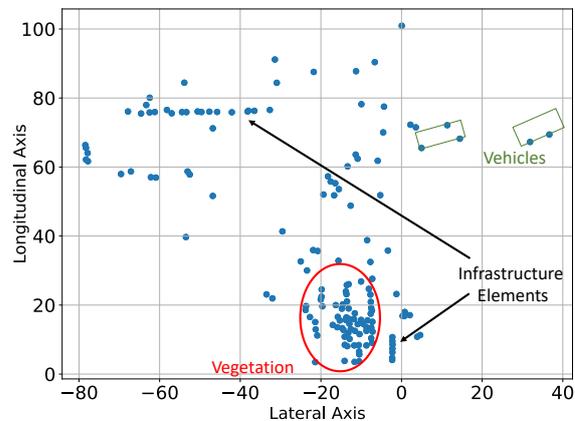

**Figure 1** A scene generated using the proposed model. The interpreted structures in the point cloud are marked.

The recent advancements in research of generative methods have opened the path for synthetically generating high-fidelity data instances [4]. Similarly, using conditional generation, even such data instances can be generated for automotive radar datasets, which rarely occur in nature, like a deer intersecting the road. We have taken inspiration from the applications of generative methods in the image and text domain to extend the research to radar scene generation using Generative Adversarial Networks (GAN).

In this paper, a method to generate radar data for a road scene using GAN architecture is proposed; an example scene is shown in **Figure 1**. The method uses a generator to capture high-level and fine features. Moreover, it takes advantage of a global discriminator to classify the whole scene, as well as a set of local discriminators, to separately classify the segments of the scene as real or fake. This set of discriminators helps in generating high-fidelity radar point cloud data. This GAN architecture has been implemented based on the point-based neural network architecture PointNet++ [5], instead of a CNN-based architecture, since the radar point cloud was used as the training set.

The GAN model architecture is described in Section 3, and the experiment setup, details, and results are discussed in

Section 4 followed by the summary and conclusion in Section 5. Before diving deeper into these topics, an overview of the related work is provided in Section 2.

## 2 Related Work

In this section, we summarize different methods to process radar data using deep neural networks to compare the advantages and disadvantages of these methods and their input level. Object detection and classification of images using neural networks have been widely popular among the research community. Methods based on convolutional neural networks are useable in the radar processing domain as well, since these methods fit well for radar 2D range-Doppler maps or range-azimuth maps [1] being used as input. Looking beyond typical applications of CNNs in radar data (target detection and classification), radar processing tasks range from object tracking to semantic segmentation of a whole scene. For these tasks, a certain pre-processing is required before feeding data into a deep neural network. So, radar point clouds are better suited as input for such networks than a 3D radar cube (or its 2D slices). Since radar point clouds are a sparse data representation, 2D convolutional neural networks are slow and perform sub-optimally on such data, hence it brings alternative network architectures based on point-based networks into focus. In [6], an approach for object detection on radar data based on graph convolutional networks has been introduced and according to the authors, it provides a 10% average precision advantage over grid-based convolutional networks. In [7] and [8], methods for semantic segmentation and tracking, respectively, using PointNet++ have been introduced.

Since the emergence of image synthesis [9], image-to-image translation [10], and text-to-image synthesis [11] using generative adversarial networks, their wide-scale applications in other domains have also come into focus. In the automotive radar domain, prominent work using GANs includes target super-resolution [12], track image generation [13], range-azimuth map generation [14], and range-Doppler map generation [15].

Despite a thorough literature review, we have not found any comparable method to generate full radar scenes synthetically, although, there are some methods to generate a single target (a vehicle, a bicycle, or a person) in a constrained environment [16],[17].

## 3 Model Architecture

The PointNet neural network architecture proposed in [18] takes a point cloud as input and performs tasks like object classification and semantic segmentation directly on the point cloud. However, since the PointNet architecture extracts the features at two levels only (per-point features and global features), it cannot capture features at different scales and generalize learning to different data densities (except the training set density). The solution to these limitations was proposed with the PointNet++ architecture [5] by feature extraction using a hierarchical point set structure consisting of set abstraction layers and feature propagation layers.

In the proposed model, we use the constituent elements of PointNet++: a feature propagation structure and a set abstraction structure embedded into a GAN architecture, as shown in **Figure 2**, to generate radar point cloud scenes. The generator consists of a *Feature Propagation* (FP) network, which takes the noise vector as input and interpolates the points in subsequent layers only using an approximation of the nearest points. Since there is no skip connection from the set abstraction layers available in the generator, the density of the point cloud is only controlled by discriminator loss. However, the maximum number of points in each feature propagation layer is controlled to guarantee convergence. The interpolated points undergo a *unit pointnet* operation, the same as a 1x1 convolution in CNNs, in each feature propagation layer. The number of feature propagation layers is set to the same as the number of set abstraction layers to obtain the same point density as in the training dataset.

The discriminator of the GAN architecture consists of hierarchical feature learning *Set Abstraction* (SA) layers, in which each layer takes a set of points as input and produces an abstracted new set of fewer elements. Each set abstraction layer is made up of three steps: a *Farthest Point Sampling* (FPS) step to select a subset of points from the input set of points, a *Multi-Scale Grouping* (MSG) step to construct local region sets of neighboring points from the sampled subset of points and a PointNet step to encode the patterns from grouped regions into feature vectors. After feature learning, a network of fully connected layers is used for the classification of the scene as real or fake. Even though, the authors in [5] deem MSG to be computationally expensive and suggest a *Multi-Resolution Grouping* (MRG) concept, we consider MSG still to be suitable for radar data since the radar data is usually much sparser than the data modes under experiment in the original paper.

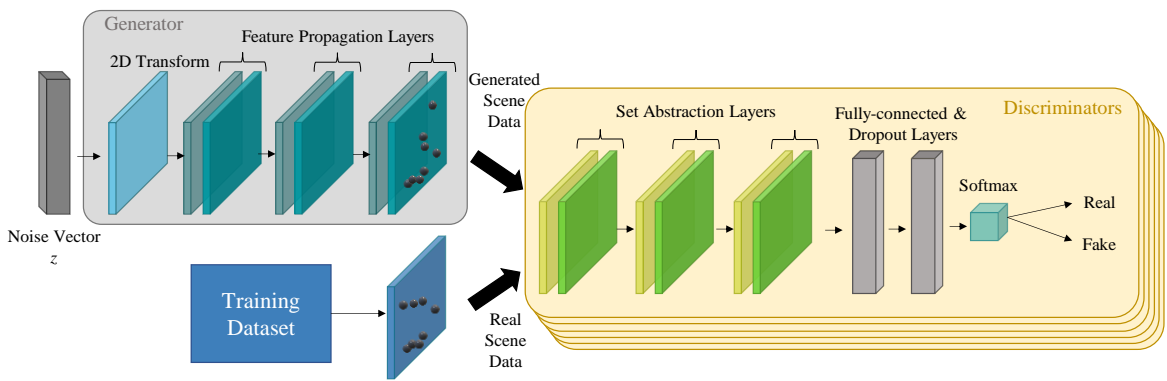

**Figure 2** Proposed model architecture including the generator and a set of discriminators (whole network and segments)

In addition to a single discriminator for the classification of the whole scene, we use six segment-wise discriminators as well, one for the each of left and right sides of the scene in near-range, mid-range, and far-range, to classify each of the segments (**Figure 3**). This helps to avoid the concentration of points in one or some of the segments.

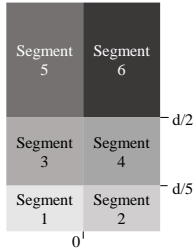

**Figure 3** Representation of scene segments used for segment-wise discriminators, where d is detection range

The total GAN loss consists of the loss of the discriminator of the whole scene and a sum of the discriminator losses of each segment weighted with λ. The resulting objective function can be expressed as:

$$\mathcal{L} = \mathcal{L}_{GAN}(G, D_{whole}) + \lambda \sum_{i=1}^{n} \mathcal{L}_{GAN}(G, D_i) \quad (1)$$

where n is the number of segments and λ is the weighting of segment-wise loss functions.

If the weighting of segment-wise discriminators (λ) is set too high then it can lead to generated mini-scenes within the segment without continuity at the edge of the segment. So, the value of λ must be less than 1 to give more weight to the discriminator of the whole scene. The number of radar detection points at closer ranges is typically higher than the farther ranges, so the segment length is set unequal in longitudinal dimension. The nearest segments span only 20% of the Field of View (FoV), the next segments span 30% of the FoV and the farther segments span the rest of the FoV (50%).

## 4 Experiments and Results

### 4.1 Dataset

The proposed network is trained on the RadarScenes [19] dataset. The dataset includes data from four radar sensors mounted on the corners of the front bumper of the vehicle. Sensors 1 and 2 are mounted at 85° and 25° horizontal tilts clockwise from the axis pointing towards the vehicle driving direction, respectively. Similarly, sensors 3 and 4 are mounted symmetrically mirrored at angles 25° and 85° counterclockwise from the axis pointing toward the driving direction.

For the model training, we used data from sensors 2 and 3 only and treated a single data sample from one sensor as a training sample without concatenation of samples from the other sensor. We applied mirroring (reverse the sign of y-axis value for each detection - axis lateral to driving direction) on the original data samples as a data augmentation technique to remove the bias caused by the mounting tilt direction. The final training set consisted of 346,633 unique original samples from the dataset and the same number of mirrored samples; similarly, we reserved 69,177 unique data samples as the validation set and 1300 data samples as the final test set (further details on the usage of the test set in Section 4.3).

### 4.2 Implementation

Except for the final layers of the generator and discriminator, we used *batch normalization* followed by an activation layer in all GAN layers (*ReLU* -rectified linear units, for generator layers, and *leakyReLU* for discriminator layers) and used *softmax* activation for the final discriminator layer. We used the Adam optimizer with the following hyperparameters: $\beta_1$=0.5, $\beta_2$=0.99 and $\alpha=2 \times 10^{-4}$. The batch size was set to 64, the value of λ from (1) was set to 0.6 and segment size was chosen for d=100 m. We implemented the model using the building blocks from [20] in PyTorch 2.0.1 and used Nvidia V100 and A100 GPUs for the training.

### 4.3 Evaluation

The evaluation of a radar scene generator (example output in Figure 1) is tricky as its visual analysis is ambiguous and reference ground truth generation is not possible. However, we used the whole scene discriminator from the model network as a real/fake classifier (mentioned as classifier from here onwards) to evaluate the generated data. We evaluated four test sets, each with 1300 samples, on the classifier: a *Real Scenes (Real)* test set consisting of 10 random samples from each training sequence (130 sequences), that were separated from the training set, and the discriminator was never exposed to these samples during training. We similarly generated the *Synthetic Scenes (Gen)* test set using the GAN model. To test our hypothesis that the synthetically generated scenes resemble the real scenes in the dataset, we evaluated two further test sets: a *Random Points (Rand)* test set of completely random data with a maximum number of detections in the scene limited to 512 and a *Curated Random Scenes (CuRand)* test set to emulate a typical radar scene with the higher density of data closer to the center of lateral axis in the FoV. We compare these four test sets based on the ratio of test samples classified as real scenes by the classifier.

### 4.4 Results and Ablation Studies

The results in row 1 of **Table 1** show that both the real scenes and synthetic scenes test set have similar performance (~87%) on the classifier, while the two sets of random data samples perform significantly poorly in comparison. We introduce another network with slight modification from the proposed network as *Ablation Model 1*, where the feature propagation and set abstraction layers consist of a single layer each (built upon the PointNet [23] architecture). The comparison of the results between this ablation model (row 2 of Table 1) and the original model highlights the advantage yielded by the sampling and grouping in the original model and that the depth of the network in this abl-ation model is insufficient to learn the features in a radar point cloud scene.

| Ratio of Data Classified as Real | | | | |
|---|---|---|---|---|
| | Test Set Type | | | |
| Model | Real | Gen | Rand | CuRand |
| PointNet++ GAN Model *(Ours)* | 0.868 | 0.875 | 0.057 | 0.134 |
| Ablation Model 1 *(single SA/FP layer)* | 0.865 | 0.205 | 0.058 | 0.136 |
| Ablation Model 2 *(one discriminator)* | 0.867 | 0.44 | 0.058 | 0.135 |
| Ablation Model 3 *(filtered training dataset)* | 0.889 | 0.883 | 0.049 | 0.104 |

**Table 1** Binary (real or fake) classifier result for the proposed model and ablation models on four test sets

We additionally test a variant of the model with only one discriminator, the results of this *Ablation Model 2* can be seen in row 3 of Table 1. From comparison to row 1 in Table 1 follows that the set of discriminators in the original model gives an improvement of ~42% on the classifier. Moreover, we came up with the hypothesis that the input data scenes with a relatively low number of detections reduce the learning ability of the model, so we filtered out the scenes with less than 30 detections from the training set, validation set, and test set, to understand its impact on classification performance. The results of this *Ablation Model 3* in row 4 of Table 1 support the hypothesis and the filtered dataset provides even better results than the results achieved using the whole dataset.

## 5  Conclusion

We have proposed a method to generate full radar scene data using a PointNet++ based GAN model architecture and showed that the generated scenes from our model perform very well on the real/fake classifier. We also demonstrated using the ablation study and comparison with the originally proposed model that the depth of the model network is critical to realistic radar point cloud scene synthesis. Additionally, we have shown through ablation studies that the segment-wise discriminators and the filtered dataset improve the ability of the network to generate high-fidelity scenes.

In future work, we plan to extend to conditional generation of scenes with desired types of classes and the number of particular instances in the scene. Further, the generalization of this scene generation across different radar datasets would also be a topic of interest in the domain.

**Acknowledgments:** The research was supported by Hightech Agenda Bayern and by the German Research Foundation (Deutsche Forschungsgemeinschaft, DFG) through the project "Scenario-based tool chain for virtual verification and validation of automotive radar" (project number 503852364). Simulations were performed with computing resources granted by RWTH Aachen University under project rwth1216.